\documentclass[conference]{IEEEtran}
\IEEEoverridecommandlockouts
\usepackage{cite}
\usepackage{amsmath,amssymb,amsfonts}
\usepackage{algorithmic}
\usepackage{graphicx}
\usepackage{textcomp}
\usepackage{xcolor}
\usepackage{subfig}
\def\BibTeX{{\rm B\kern-.05em{\sc i\kern-.025em b}\kern-.08em
    T\kern-.1667em\lower.7ex\hbox{E}\kern-.125emX}}
\begin{document}

\title{Driver Drowsiness Classification Based on Eye Blink and Head Movement Features Using the \textit{k}-NN Algorithm}

\author{\author{\IEEEauthorblockN{
			Mariella Drei\ss{}ig\IEEEauthorrefmark{1}\,{\includegraphics[scale=0.075]{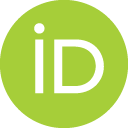} 0000-0002-8983-7514},
			Mohamed Hedi Baccour\IEEEauthorrefmark{1},\\
			Tim Sch\"ack\IEEEauthorrefmark{1}{\includegraphics[scale=0.075]{images/orcidlogo.png} 0000-0002-1184-3172} and
			Enkelejda Kasneci\IEEEauthorrefmark{3}\,{\includegraphics[scale=0.075]{images/orcidlogo.png} 0000-0003-3146-4484}}
		\IEEEauthorblockA{\IEEEauthorrefmark{1}Mercedes-Benz AG, Mercedes-Benz Technology Center, Sindelfingen, Germany}
		\IEEEauthorblockA{\IEEEauthorrefmark{3}Human-Computer Interaction, University of T\"ubingen, T\"ubingen, Germany\\
			\textit{\{mariella.dreissig, mohamed\_hedi.baccour, tim.schaeck\}@daimler.com}, \textit{enkelejda.kasneci@uni-tuebingen.de}}}
}

\maketitle

\begin{abstract}
Modern advanced driver-assistance systems analyze the driving performance to gather information about the driver's state. Such systems are able, for example, to detect signs of drowsiness by evaluating the steering or lane keeping behavior and to alert the driver when the drowsiness state reaches a critical level. However, these kinds of systems have no access to direct cues about the driver's state. Hence, the aim of this work is to extend the driver drowsiness detection in vehicles using signals of a driver monitoring camera. For this purpose, 35 features related to the driver's eye blinking behavior and head movements are extracted in driving simulator experiments. Based on that large dataset, we developed and evaluated a feature selection method based on the \textit{k}-Nearest Neighbor algorithm for the driver's state classification. A concluding analysis of the best performing feature sets yields valuable insights about the influence of drowsiness on the driver's blink behavior and head movements. These findings will help in the future development of robust and reliable driver drowsiness monitoring systems to prevent fatigue-induced accidents.
\end{abstract}


\section{Introduction}
Drowsy driving is a controversial topic when coming to road safety. Nearly everyone who drives a car on a regular basis already experienced drowsiness or even micro-sleeps during driving. Yet it is a topic with a fairly low awareness in society. Nevertheless, throughout the years 2008 to 2018 the frequency of drowsiness-induced accidents in Germany increased \cite{Statista2019}.

That indicates a higher need for reliable drowsiness monitoring systems in vehicles. Major functions of such a system are to assist the driver to better assess drowsiness and to prevent severe impairments of the driving skills.\par

A driver drowsiness monitoring system can be based upon different measures around the vehicle and/or the driver. Some of the driver drowsiness monitoring approaches aim to build a system on one single measure, while the majority of the modern approaches in fact rely on a combination of measures (so-called hybrid methods). This is particularly beneficial in complex real-world scenarios where a single measure might not catch the driver's state sufficiently. Thus, the detections can be validated with additional information from other domains, increasing the drowsiness classification confidence \cite{Dong2011}. \par 

Nevertheless, it is a prerequisite to thoroughly understand the distinct features indicating the driver's level of drowsiness. The aim of this work is to estimate the driver's state based on behavioral measures, namely the head movement and blink features of drowsy drivers, and recommend a break in the case that certain signs of sleepiness are detected. Another purpose of this work is to gain insights about certain behavioral characteristic to enable further development of robust and reliable driver state classification systems. For this purpose, the $k$-Nearest Neighbor ($k$-NN) algorithm is used to classify the driver's state of drowsiness based on the eye closure and head movement characteristics. \par

This paper is organized as follows: Section \ref{sec:literature} gives an overview on the state-of-the-art techniques on driver drowsiness detection and state classification. Section \ref{sec:data} summarizes the data and features used in this work, including a description of the the data collection process and the feature extraction and analysis. Section \ref{sec:classification} presents the core points of the $k$-NN based driver drowsiness state classification: the treatment of the classification problem, the model design with the search for a suitable distance metric and value of $k$, the feature selection part based on the previously designed models and ultimately suggestions about the model refinement. Sections \ref{sec:discussion} and \ref{sec:conclusion} discuss and conclude the paper briefly.

\section{Drowsiness Detection Techniques and Basic Principles} \label{sec:literature}
Several car manufacturers offer driver assistance systems that estimate the driver's state and recommend actions accordingly. Despite promising advances in the research and development of driver drowsiness detection systems, further investigations are required to improve their performance. By that, crucial steps towards an increased road safety are taken.

\subsection{Measures for Driver Drowsiness Detection}
Drowsiness comes along with cognitive impairments that are particularly dangerous during driving. Decades of research have identified measures that enable a driver state determination. They can be divided into three categories: driver behavior, physiological responses and driving performance \cite{Dong2011}. \par

The behavior of a drowsy driver shows phenomena like increased frequencies of head nodding, yawning and eye blinking as well as more frequent body movements in general (like scratching the head and face or moving the legs) \cite{Eskandarian2007}. For example, the authors of \cite{Friedrichs2010} utilize several blink-related features for classifying the driver's state of drowsiness. \par

In addition to a notable modification of the behavior, drowsy humans exhibit several altered physiological responses. Signs of increasing drowsiness can be found in electrooculography (EOG) \cite{Hu2009}, electroencephalography (EEG) \cite{Jap2009} and electrocardiography (ECG) \cite{Jung2014} signals. For example, the authors of \cite{Sommer2010} compared three approaches to drowsiness detection: video-based, EEG/EOG-based and lane keeping-based techniques. Their findings show that EEG and EOG indicate driver drowsiness better than the video-based PERCLOS feature. Nevertheless, the invasive nature of those measures renders it impossible to install them in a vehicle for a daily application.\par

Vehicle-based drowsiness indicators stem from the most critical trait of drowsy drivers: an impaired driving performance. This decline in driving ability is mainly a result of increased reaction times \cite{Philip2003}. The approach by \cite{Friedrichs2010a} relies on a set of steering and lane keeping features to classify a driver's state of drowsiness. Other methods aim to further improve established vehicle-based driver drowsiness detection methods and increase its robustness in real-world applications \cite{Spindler2015}.

Recent works on driver drowsiness detection and classifiation combine multiple modalities to increase the algorithm's reliability and robustness. For example, the authors of \cite{JacobedeNaurois2018} include behavioral and physiological data into their work. The authors of \cite{Daza2014} build their system on behavioral and vehicle-based drowsiness indicators. The approach presented in \cite{Gwak2018} even combines different features from all three domains.\par

For a further thorough review on recent advances in driver drowsiness detection systems, see \cite{Sikander2019}. 

\subsection{Drowsiness Classification Methodology}
Many driver drowsiness state classification approaches are implemented via knowledge-driven fuzzy systems \cite{Azim2014} or statistical analysis \cite{Cheng2012}. Of particular interest in this context are data-driven machine learning models, which can be divided into conventional, shallow models like Support Vector Machines or the $k$-NN method, \cite{Gwak2018, Li2017} and deep models like (convolutional) Artificial Neural Networks \cite{JacobedeNaurois2018}. The basic idea of machine learning is to train an algorithm on an extensive database, with the aim that the algorithm finds an accurate representation of the underlying distributions by itself. \par

We used the $k$-NN method \cite{Cover1967, Silverman1989} for driver state classification. To the best of our knowledge, it is not previously studied in the context of a camera-based driver drowsiness detection using blink features. Existing $k$-NN-based approaches include the steering behavior \cite{Li2017}, EEG measures \cite{Jalilifard2016} or facial features \cite{Nakamura2013}. The work of \cite{Ebrahim2016} investigates the feasibility of a drowsiness classification system based on blink features gathered with an EOG. The author achieved a promising classification accuracy, indicating the potential of a $k$-NN classifier in combination with blink-based features. \par

The $k$-NN method is a so-called lazy learning technique, where the algorithm simply stores the data for later processing instead of learning a compact representation of it \cite{Aha1997}. This kind of modeling technique has some clear advantages. For example, the dataset can be simply updated by adding new datapoints. Thus, the resulting models are flexible and able to adapt to domain changes. As the generalization of the input data is delayed until new queries are made to the system, the $k$-NN algorithm does not require any training time. Furthermore, $k$-NN models are transparent and have a high explainability. That means its decisions can easily be understood and analyzed, which allows to gather additional knowledge about the effect of drowsiness on the driver's head movement and blink behavior. These advantages can be exploited even with a relatively small dataset. \par

The $k$-NN model requires a set of suitable features as basis for the classification, especially when a high-dimensional feature space is available. According to the "curse of dimensionality" phenomenon \cite{Bellman2003}, the available data becomes sparse as the number of possible configurations grows. Thus, one aim of this work is to identify a suitable set of meaningful features.

The feature selection techniques mainly used in this work are wrapper methods \cite{Kohavi1997}. Wrapper methods select feature subsets according to their predictive value in the actual classification process. Therefore, this method is capable of considering dependencies between the feature subset and the classifier as it directly assesses the classification performance. 

\section{Data and Feature Engineering} \label{sec:data}
The first step in the process of developing a camera-based classification system is to define the state of drowsiness in terms of observable measures. Especially the eyelid movements as drowsiness indicators were studied through the decades. The authors of \cite{Wierwille1994} introduced the PERCLOS measure which denotes the proportion of time in a defined interval for which the eyes are more than $80\%$ closed. The work of \cite{Schleicher2008a} hints that the blink behavior in general is an observable indicator for drowsiness. Besides the PERCLOS, other blink-related features can be extracted from an eyelid movement signal \cite{Ebrahim2016, Baccour2018}. Most of these features exhibit an altered behavior with an increasing drowsiness.

\subsection{Data Collection}
We recorded about 134 hours of material during three driving simulator studies, with the objective to obtain data that reflect the interactions between the eyelid closure and the drowsiness. Therefore, a camera facing the driver is set up to detect and track the eyelid movements. It is placed on the steering wheel column and is equipped with infrared illumination for robust eye and head tracking. \par

The camera provides several signals related to the driver's head position, gaze direction and eyelid closure. Among them, four signals are of interest for this work, including the eye closure and the eyelid confidence for the blink feature extraction and the roll and pitch angle of the \textit{head rotations}. For the eye closure, the maximal distance $Ld$ between the upper and lower eyelid was measured. The eye closure value was then computed with the following equation:
\begin{equation}
	\textrm{eye closure} = \max\left(1-\frac{Ld}{Ld_I}, 0\right),
\end{equation}
with $Ld_I$ being the iris diameter, which is assumed to be $12$mm for every subject. Thus, the eye closure signal can take values between 0 and 1, where 1 means the eye is completely closed. To avoid false detections during phases where the driver's face is (partially) occluded, e.g., by the steering wheel or head rotations, a confidence signal is provided. For further details about the signal preprocessing, we refer to \cite{Baccour2019}. \par

To assess the driver’s drowsiness level as reference, the drivers were requested to report according to the Karolinska Sleepiness Scale (KSS) \cite{Akerstedt1990} in intervals of $15$min. This nine-point scale (1 = extremely alert, 3 = alert, 5 = neither alert nor sleepy, 7 = sleepy but  no  difficulty  remaining  awake and 9 = extremely sleepy / fighting sleep) is widely used for self-reporting drowsiness levels as it is closely related to physiological drowsiness measures and thus a reliable and valid drowsiness indicator \cite{Kaida2006}. The raw KSS values are validated by a more sophisticated supervision procedure which includes adjusting them with the help of expert evaluations, PERCLOS values and lane keeping behavior (compare \cite{Svensson2004}). \par

To induce drowsiness during the driving simulator experiments, the track was designed in a circuit route (130km) and as monotonous as possible. The track resembled a two-lane German highway at night with low traffic density. The subjects were instructed to drive at the recommended speed of $120\frac{km}{h}$ and were not allowed to use advanced driver assistance systems like the Adaptive Cruise Control. The duration of the experiment depended on the subjective drowsiness reporting and was on average $103\pm37$min over all experiments. Further details on the experimental setup are provided in \cite{Baccour2018}. 

\subsection{Feature Extraction and Analysis}
In \cite{Baccour2018} the author proposed a robust technique to detect blinks and derive blink features based on the eye closure signal. Utilizing this method, a list of 35 blink features build the base of this work. They can be grouped into eleven categories, depending on which domain they stem from: frequency, time, amplitude, velocity, amplitude-velocity ratio, percentage, blink (form), eyelid, head movement, symmetry and the PERCLOS\footnote{Two variants of the PERCLOS measure were adopted in this work. In the following, PERCLOS1 denotes the mean proportion of a time window where the eye is more than 80\% closed, whereas PERCLOS2 divides that value by the blink duration in the same time window.}. \par

With regard to the signal processing and feature extraction, it has to be taken into consideration that drowsiness does not occur suddenly but rather has a slow progression. To be able to observe the adaption trend over a longer period of time, the eye closure signal is processed in a sliding window manner. Within a window width of $\Delta T=10$min, the mean and standard derivation (std) of the features are calculated to represent the compressed values of that window. The window is then moved with a step size of $h=1$min. \par

A sophisticated driver drowsiness classifier is supposed to handle interindividual differences properly and achieve high classification accuracies on all subjects equally. Yet, differences between the individuals blinking behavior can be tremendous. To cope with that, a baselining of the features is introduced by defining the first $10$min of each experiment as a representation of the \textit{awake} state for each subject (see \cite{Friedrichs2010}). This underlies the assumption that the subject is still moderately awake when starting the experiment. Drivers who started the experiment in an already drowsy state are at a high risk of injecting invalid data into the underlying dataset and thus distort the model. Therefore, these subjects are excluded in advance from the dataset.

\section{Driver Drowsiness Classification with $k$-Nearest Neighbors} \label{sec:classification}
The driver state classification consists of two main parts: the model design and the feature selection with wrapper methods. The model design step includes setting all the framework conditions for the optimization process in the feature selection step. But prior to that, some preprocessing steps are necessary.

\subsection{The Classification Problem}
As explained previously, the output of the classification model should be an estimation whether the driver should be warned, possibly warned or not warned. Therefore, the KSS labels have to be converted into the classes \textit{awake} (do not warn), \textit{questionable} (possibly warn) and \textit{drowsy} (warn). The definition of the classes highly depends on the wanted warning behavior of the driver drowsiness detection system. It is important to find a good trade-off between a sensitive system which warns at the sight of first drowsiness indicators and a rather cautious system which warns only in extreme cases. For this work the binning proposed in \cite{Ebrahim2016} is applied, resulting in the following class distributions:
\begin{itemize}
	\item binary classification
	\begin{itemize}
		\item KSS $\in [1,6]$: \textit{awake} (45\%)
		\item KSS $\in [7,9]$: \textit{drowsy} (55\%)
	\end{itemize}
	\item multiclass classification
	\begin{itemize}
		\item KSS $\in [1,5]$: \textit{awake} (31\%)
		\item KSS $\in [6,7]$: \textit{questionable} (36\%)
		\item KSS $\in [8,9]$: \textit{drowsy} (33\%)
	\end{itemize}
\end{itemize}

\subsection{Model Design}
The aim of the model design process is to construct a framework that enables an optimal search for feature set that is suitable for the classification task. Most important is the choice of $k$, which defines the neighborhood in the $k$-NN method.\par

\subsubsection{Preliminary Considerations and Preparations}
Prior to the search for a suitable value of $k$, a proper distance metric is needed to define the $k$ nearest neighbors of a datapoint in the high-dimensional feature space. The definition of the similarity $d$ between two datapoints and therefore of the neighborhood highly influences the performance of the $k$-NN classifier. In \cite{Cover1967} the authors stated that the $k$-NN algorithm works with every distance metric but in practice some metrics turned out to be particularly useful \cite{Williams2010}. Among them are the Minkowski distance, more specifically the \textit{Euclidean distance}. \par

In order to consider all features equally when computing the Euclidean distance, the data have to be scaled properly, e.g., by the z-score standardization (zero-mean and unit-variance). In this work, a slight adaption of this scaling method is applied to avoid distortion caused by skew data due to drowsiness. In other words, the data gathered during drowsy phases are not seen as a representative behavior and are excluded when calculating the population's mean and standard deviation \cite{Simon2013}. \par 

The classification results are validated with leave-one-subject-out cross-validation due to interindividual differences and the fact that the features encode a chronological trend. In this cross-validation scheme with $N_{ds}$ folds, where $N_{ds}$ corresponds to the total number of driving sessions, the model is fittet using the samples of $N_{ds}$-1 driving sessions and evaluated on the samples of the left-out driving session (unseen data). Thus, this cross-validation scheme ensures that the model generalizes well on future unseen data. 

\subsubsection{Search for a suitable $k$}
Most crucial for the failure or success of a $k$-NN classifier is the choice of the parameter $k$. Decades of research were spent on finding rules for determining the optimal value for $k$ \cite{Hall2008}. It has to be noted that there is no universal answer on the question for the best $k$ \cite{Cover1967}. The choice of $k$ is rather an optimization problem with the aim of finding the beneficial trade-off between two objectives. Small values for $k$ can lead to a noisy data representation which is prone to a distortion by outliers. On the other hand a larger $k$ gives smoother decision boundaries which means a lower variance within the defined classes \cite{Cover1967}. \par

For these reasons a range of possible $k$ will be determined in this work by analyzing the classification performance for $k$ being in the interval $[1,1000]$ (the upper boundary of $k = 1000$ is chosen based on the number of available data points in the dataset described above). The following feature selection will be conducted on every suitable candidate $k$ to find the individual optimal $k$ for each setting. This approach has the additional advantage that interactions between the value of $k$, the feature set length $p$ and the chosen features can be revealed. \par

To assess a model's performance, suitable metrics are required which reflect the model's efficiency with respect to the given task. The aim of the driver state classification is to detect a \textit{drowsy} (positive event) and an \textit{awake} state (negative event) as often as possible. As the dataset is not perfectly balanced, the Balanced Accuracy (BA) metric is convenient for that purpose. At the same time the type I error has to be kept low. This is exceptionally important because the driver is likely to be irritated by a system that outputs false alarms \cite{Dong2011}. The acceptance towards the system will decrease when the user faces a warning that seems to be unfounded and arbitrary. Therefore, special attention is paid to the False Positive (Drowsy) Rate (FDR). \par

The search for $k$ is conducted with the help of the most promising feature set candidates identified by a filter method, specifically the correlation-based feature selection (CBFS) \cite{Hall1999}. This method is supposed to find a trade-off between the relevance and the redundancy by setting them into relation to each other and thus obtaining a joined measurement criterion that can be optimized (maximizing the relevance while minimizing the redundance). The resulting feature set is: reopening duration, eyelid cleft, head nodding, std of closing duration, PERCLOS1, PERCLOS2, mean area defined by the blink and std of opening amplitude. \par

\begin{figure}[!htb]
	\centering 
	\subfloat[Performance trend of the binary classification for different $k$'s on the feature set defined by the correlation-based feature selection] {{\includegraphics[width=0.9\columnwidth]{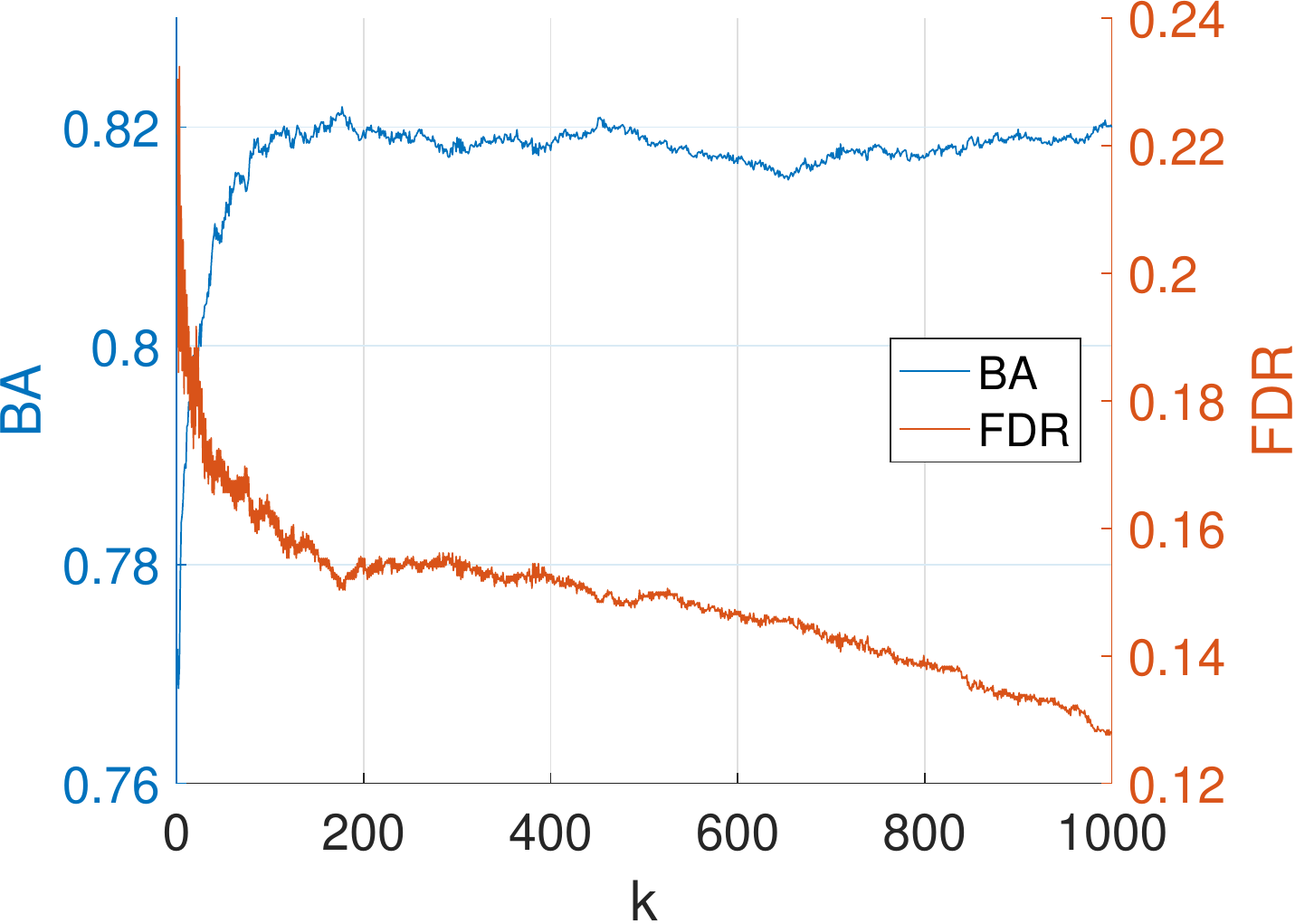}}} \\
	\subfloat[Performance trend of the multiclass classification for different $k$'s on the feature set defined by the correlation-based feature selection] {{\includegraphics[width=0.9\columnwidth]{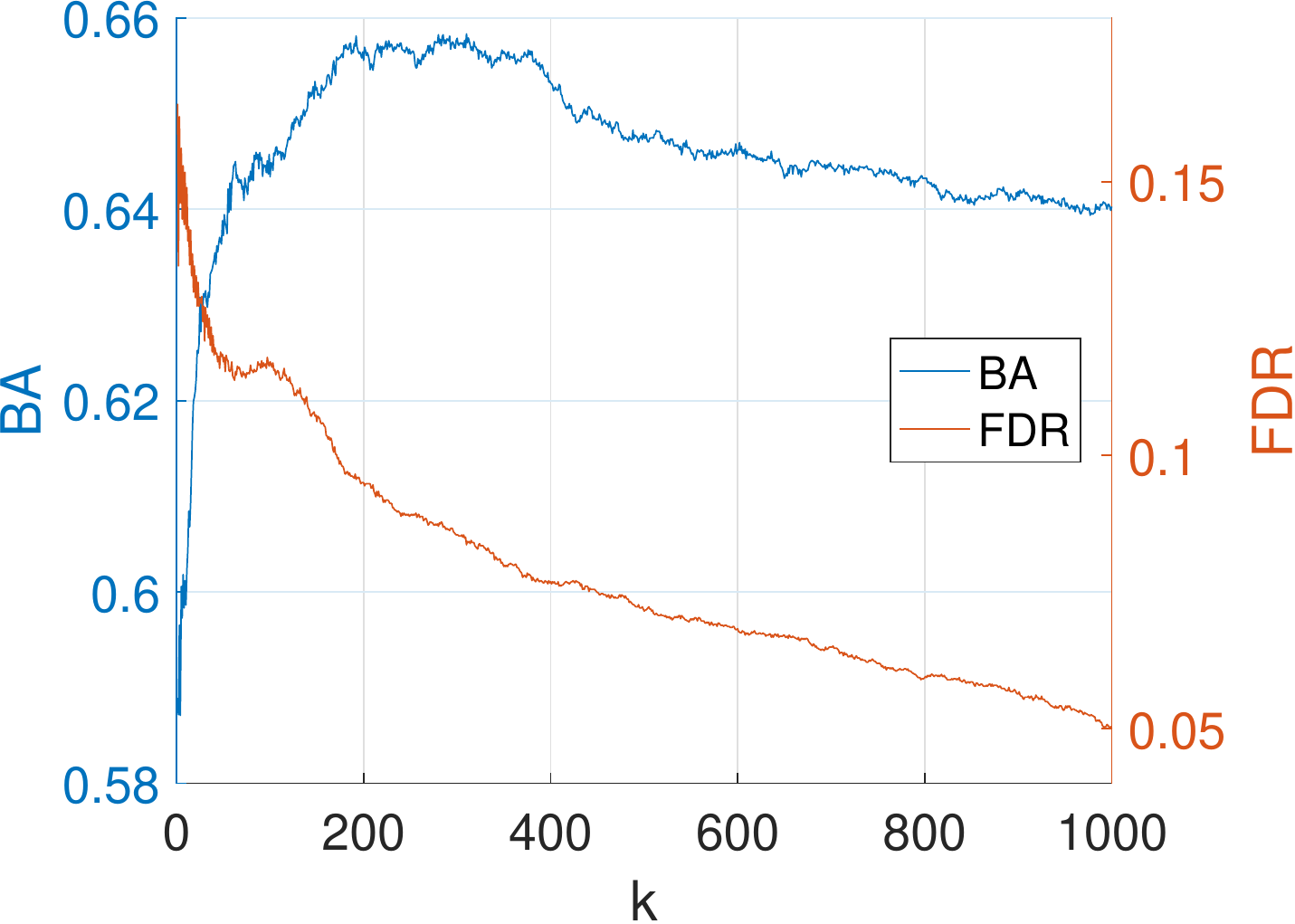}}} 
	\caption{Search for a suitable $k$ for the binary and the multiclass driver drowsiness state classification.} \label{fig:search_for_k}
\end{figure}

Fig. \ref{fig:search_for_k} shows the resulting BA's and FDR's for $k \in [1,1000]$ in both binary and multiclass classification setting using the features selected by the CBFS method. 

As can be seen, the BA initially rises with increasing $k$ whereas the FDR decreases. In both settings the performance metrics exhibit a noticeable optimum at $k \in [50; 400]$ (maximum BA and minimum FDR). After that, the BA curves either drop or stagnate, indicating no further improvement of the model's performance. Therefore, in the following feature selection procedures with wrapper methods, values for $k$ between $50$ and $400$ will be tested. To further reduce the computational costs when searching for $k$, the step size of $k$ is set to $25$. 

\subsection{Feature Selection}
After setting an appropriate distance metric and suitable values for the parameter $k$, wrapper methods are applied to select the most discriminative features for the classification. An exhaustive search for the optimal feature set is not feasible when the feature set length $p$ becomes too large. Feature subset selection methods rely on computationally feasible procedures which may yield sub-optimal results \cite{Somol2010a}. All classification results are again validated with leave-one-subject-out cross-validation to ensure the generalization ability of the model.

\subsubsection{Sequential Feature Selection}
The sequential feature selection in general aims to construct feature sets by electing one feature at a time in a greedy manner. In each step, all candidate features are evaluated with respect to their predictive power given the already contained features. The one with the greatest contribution to the model's performance will be included in the model and cannot be a candidate again. The algorithm terminates when eventually all features are selected. \par

The Sequential Forward Selection (SFS) reduces the search space by starting with zero features and adding one feature after another. In every iteration step the feature with the best performance is selected and cannot be selected again. 

\begin{figure}
	\centering 
	\subfloat[The binary classification model's performances depending on $k$ and $p$] {{\includegraphics[width=0.9\columnwidth]{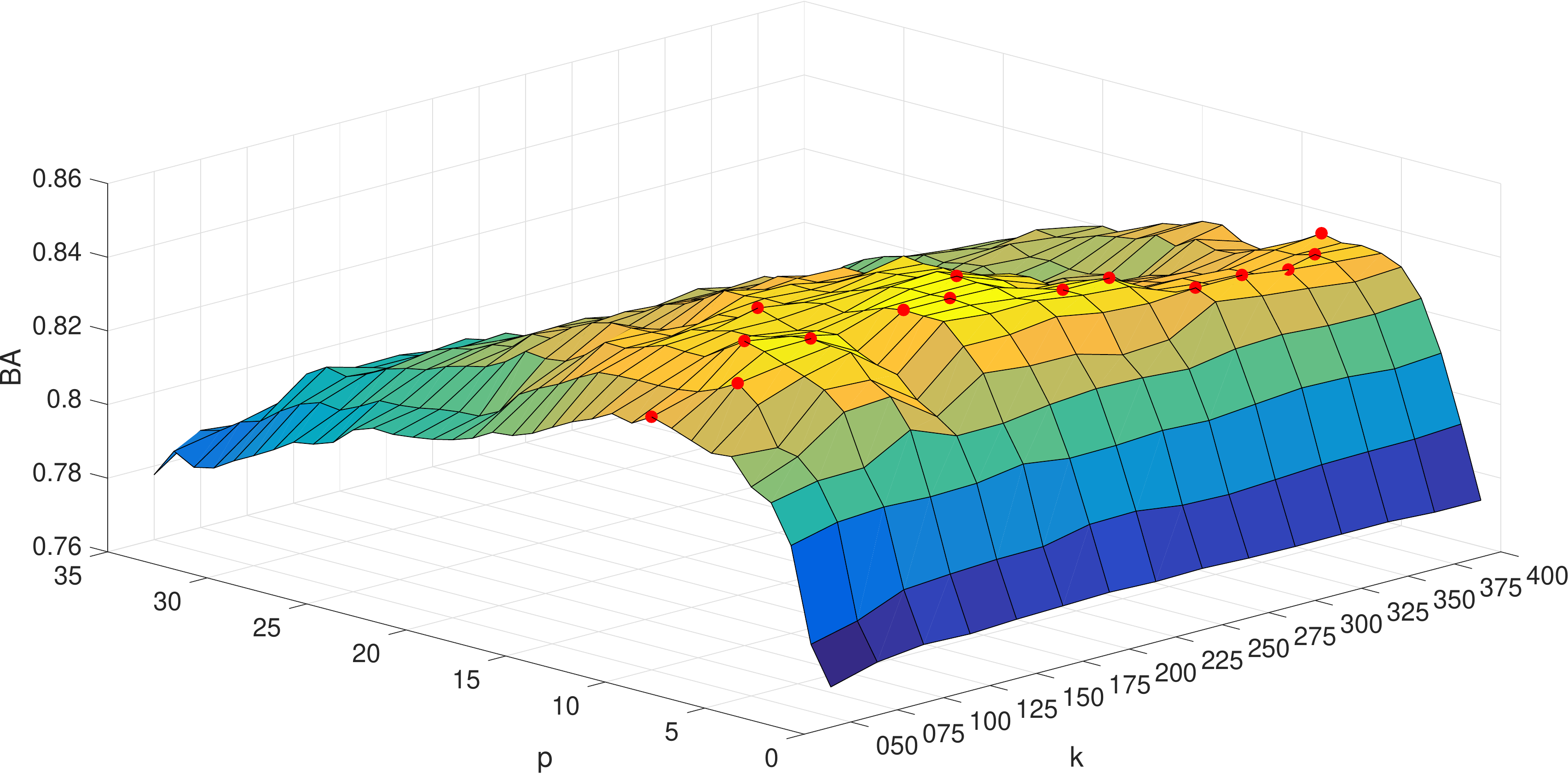}}} \\
	\subfloat[The multiclass classification model's performances depending on $k$ and $p$] {{\includegraphics[width=0.9\columnwidth]{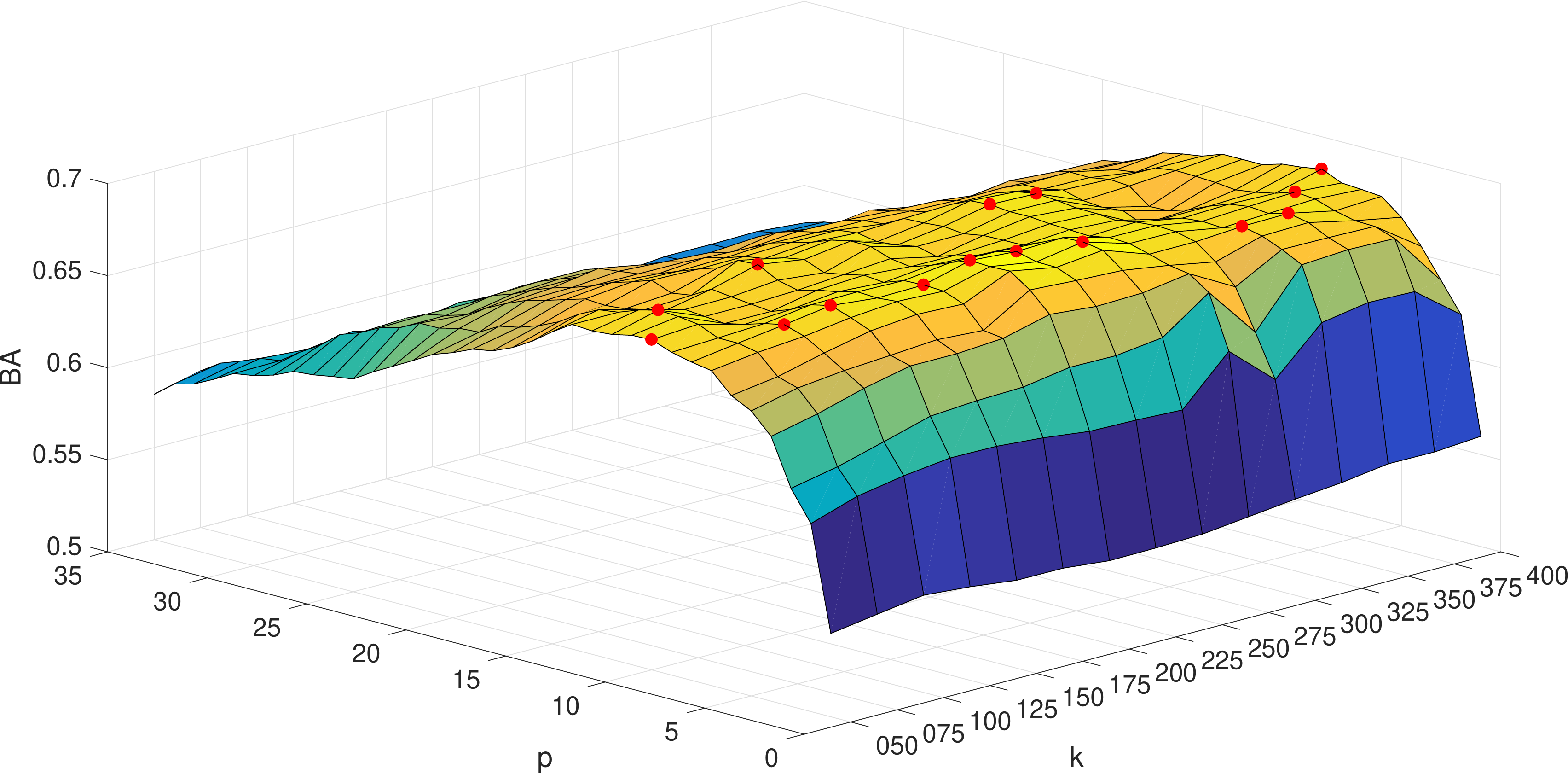}}} 
	\caption{BA trends of the classification models using the feature sets identified by the SFS method. For each $k$ the best performing model is marked with a red dot.} \label{fig:sfs}
\end{figure}
Fig. \ref{fig:sfs} shows the classification BA depending on the choice of $k$ and $p$ for the SFS algorithm. The BA rises with an increasing feature set length $p$, as the included features add to the overall classification performance. A feature set length of $p \approx 10$ maximizes the model's performance. Afterwards the BA declines again, indicating that the composed feature sets contain too much redundant features. 

\subsubsection{Sequential Floating Feature Selection}
The inherent problem of sequential feature selection is that it creates nested feature sets, no matter in which direction it is conducted. As a result, the greedily selected subsets always contain the subset of the previous step. To cope with that, floating feature selection methods introduce a flexible backtracking that allows the algorithm to dynamically find a trade-off between forward and backward steps \cite{Pudil1994}. Consequently, $p$ "floats" up and down as there is no determined feature set length per iteration step. The algorithm terminates when either a specified feature set length is reached or it got stuck in a loop by revisiting the same feature sets over again. The near optimal feature sets obtained by this technique require a higher computational time, yet it is still faster than an exhaustive search. \par

This technique can be applied in a forward (Sequential Floating Forward Selection, SFFS) and in a backward manner (Sequential Floating Backward Selection, SFBS). The SFFS method consists of two altering phases: the inclusion and the conditional exclusion. Starting with an empty feature set, the first step includes the feature with the maximum predictive power. In the second step, backward steps are performed as long as the classification performance increases. This is repeated until removing features does not longer lead to an improved performance. The SFBS method works contrary, the algorithm starts with a full feature set and successively removes one feature after another. After each step, it performs a variable number of conditional inclusion (forward) steps until the classification performance does not increase anymore. 

\begin{figure}
	\centering 
	\subfloat[The binary classification model's performances depending on $k$ and $p$] {{\includegraphics[width=0.9\columnwidth]{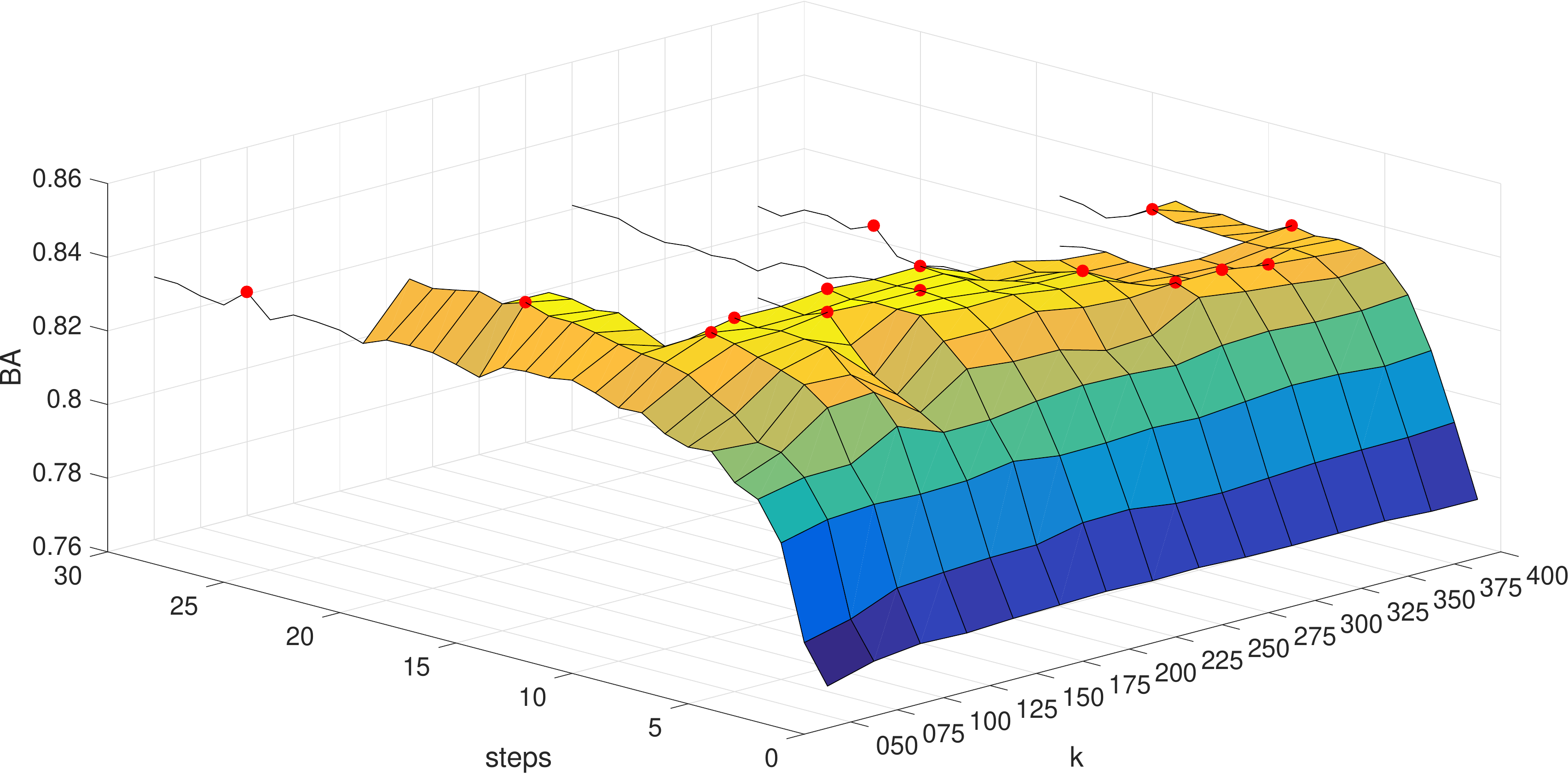}}} \\
	\subfloat[The multiclass classification model's performances depending on $k$ and $p$] {{\includegraphics[width=0.9\columnwidth]{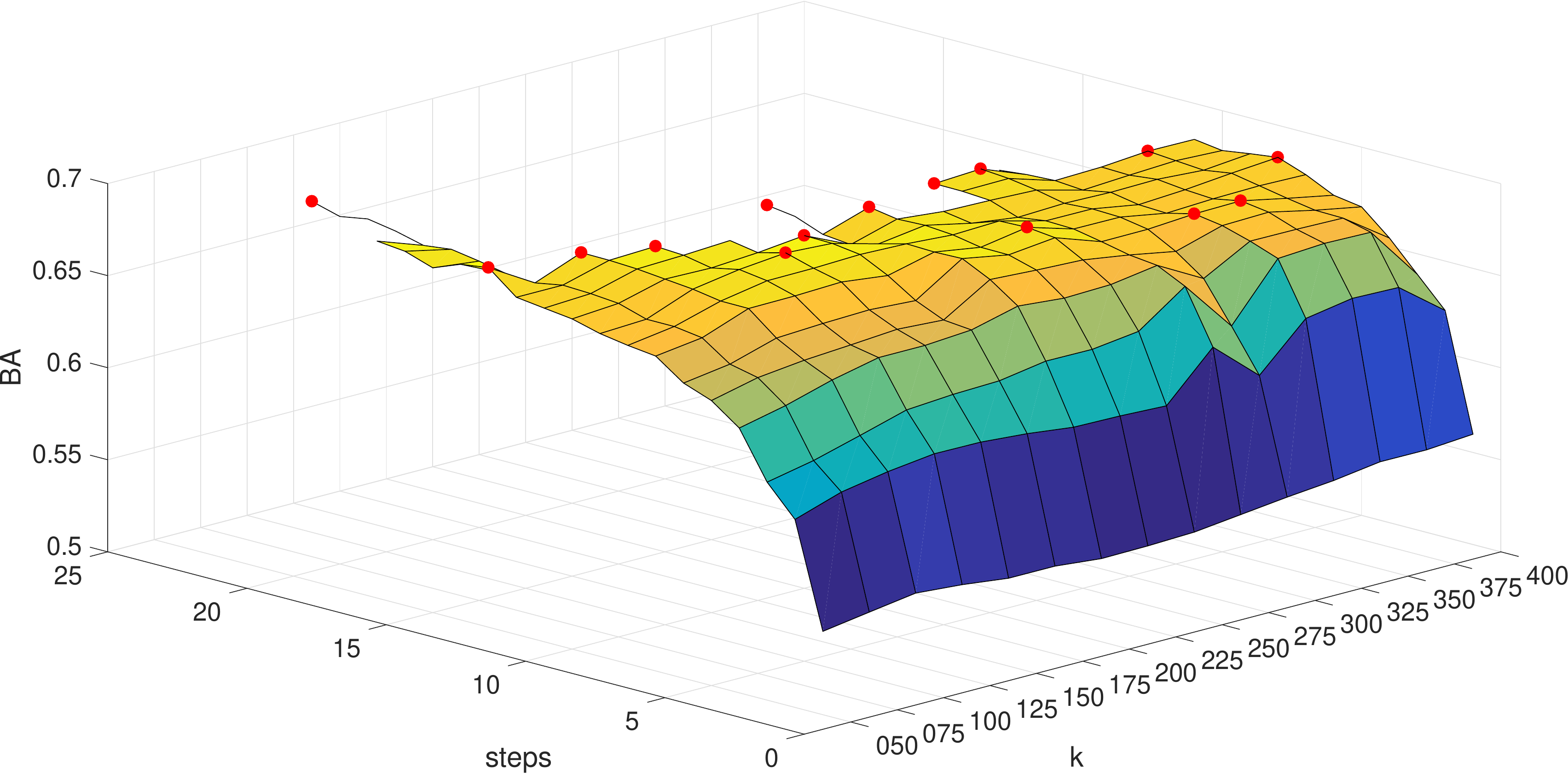}}}
	\caption{BA trends of the classification models using the feature sets identified by the SFFS method. For each $k$ the best performing model is marked with a red dot.} \label{fig:sffs}
\end{figure}
Fig. \ref{fig:sffs} shows the BA trend across the SFFS algorithm iterations for each $k$. It can be observed that due to the method's flexibility for some $k$'s the algorithm converges faster into a near optimal feature set. The number of required steps and thus the algorithm runtime is not determined in this feature selection method and thus fluctuates.

\begin{figure}
	\centering 
	\subfloat[The binary classification model's performances depending on $k$ and $p$] {{\includegraphics[width=0.9\columnwidth]{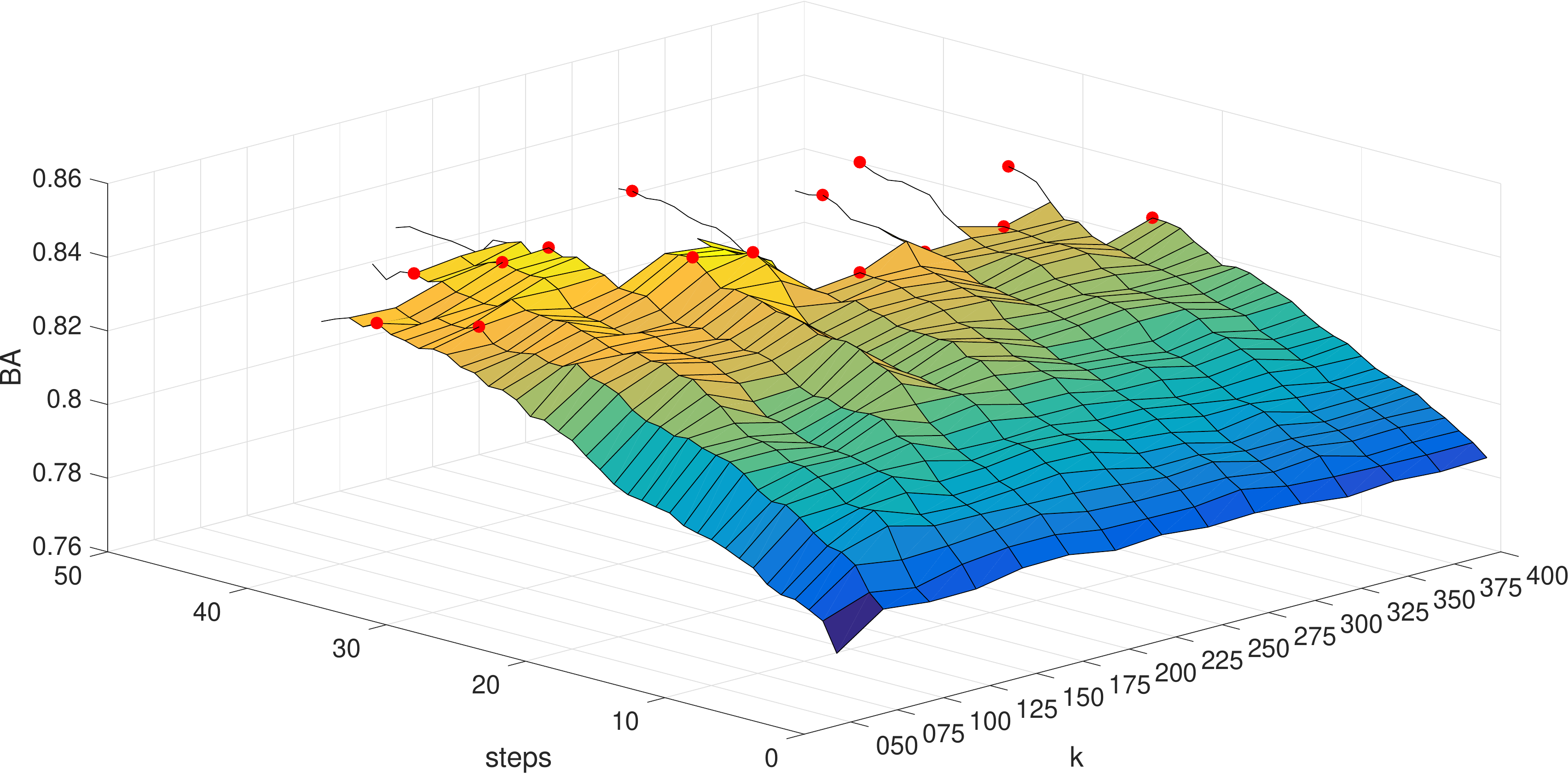}}} \\
	\subfloat[The multiclass classification model's performances depending on $k$ and $p$] {{\includegraphics[width=0.9\columnwidth]{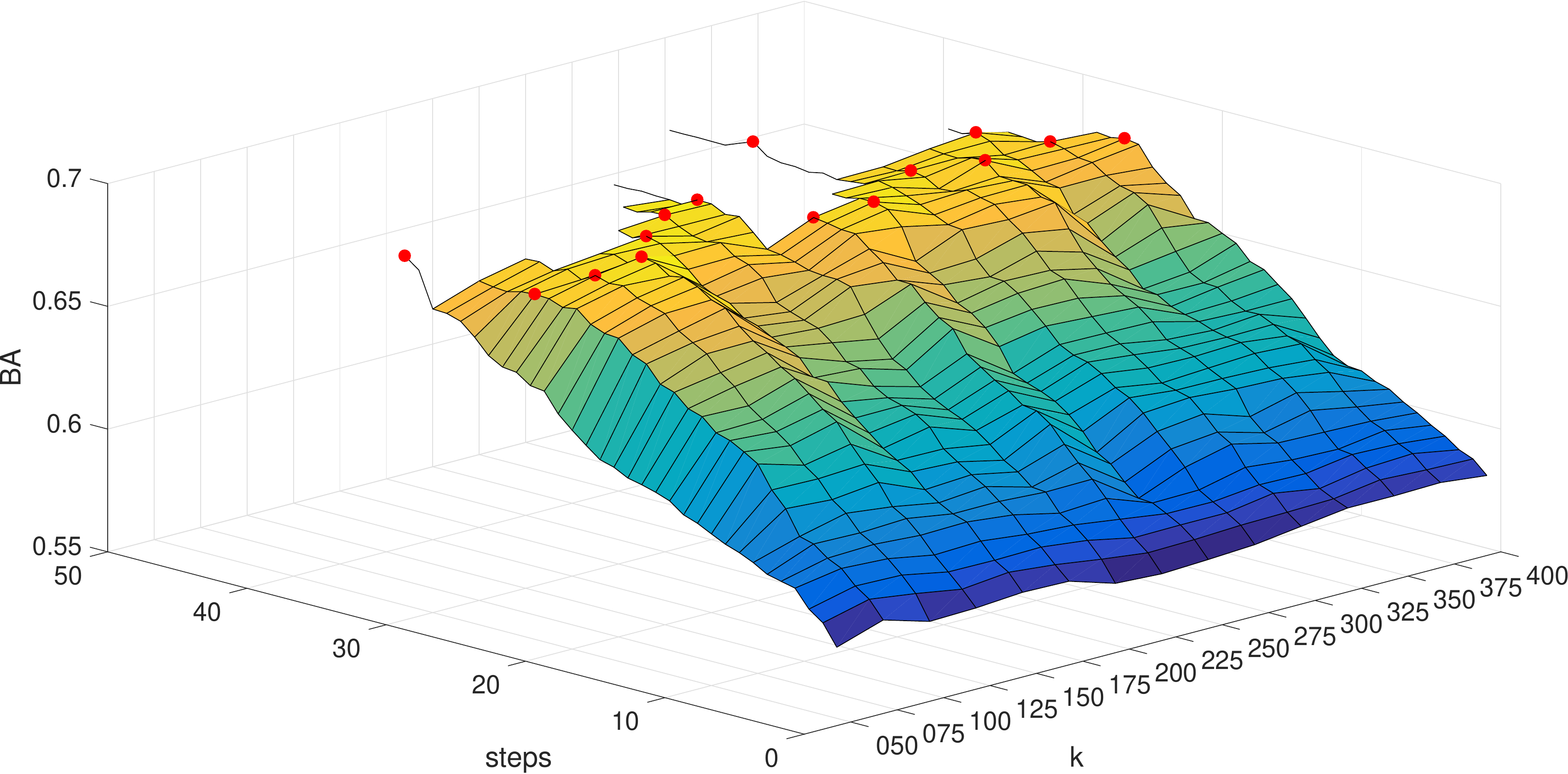}}}
	\caption{BA trends of the classification models using the feature sets identified by the SFBS method. For each $k$ the best performing model is marked with a red dot.} \label{fig:sfbs}
\end{figure}
Fig. \ref{fig:sfbs} shows the BA trend across the SFBS algorithm iterations for each $k$. Compared to SFFS, the SFBS method requires more iterations to remove the irrelevant and redundant features. Generally it exhibits a similar behavior like SFFS. 

\subsubsection{Feature Selection Evaluation}
Table \ref{tab:eval_FS} lists the best performing feature sets, the respecitve $k$'s and their performances for each feature selection approach. Some features exhibit a higher predictive power than others, indicating their usefulness for the development of a reliable and well-generalizing drowsiness detection system. The results show furthermore that the binary classification yields promising results while the multiclass classification lacks behind in terms of performance. It should be also noted that all presented feature selection approaches construct feature sets of roughly equal length. \par

Among the 35 blink features extracted from the eye closure signal, only a few are really meaningful and can be used for an efficient driver state classification. Generally, features from the category time-based are picked more frequently than features from other domains. Additionally, the percentage-based features seem to perform well in classifying a driver’s drowsiness level, as well as the head movement-based features and the eyelid cleft. The best performing feature sets identified by the different feature selection techniques suggest that some feature combinations seem to be especially important for a robust classification: reopening duration, amplitude-velocity ratio of the closing phase, PERCLOS2, eyelid cleft, head nodding and head bobbing. At the same time some other features seem to be exchangeable, so it can be assumed that they either contain redundant information or that they only contribute little to the feature set's overall classification performance. This leads to multiple feature combinations resulting in almost the same classification performance. This complicates the choice of an optimal set of blink and head movement features for driver drowsiness state classification and emphasizes the importance of defining a proper objective and performance metric. \par

\begin{table*}[t]
	\centering
	\begin{tabular}{p{1.45cm}p{0.75cm}p{0.45cm}p{11.25cm}p{0.75cm}p{0.75cm}}
		\hline \hline classification & method & $k$ & feature set & BA & FDR \\ \hline
		binary & SFS & 200 & reopening duration, eyelid cleft, amplitude-velocity ratio of the opening phase, head bobbing, head nodding, std of mean area defined by the blink, std of blink duration, blink frequency, PERCLOS2 & 0.8425 & 0.0979 \\
		& SFFS & 250 & reopening duration, eyelid cleft, amplitude-velocity ratio of the opening phase, head bobbing, std of mean area defined by the blink, head nodding, PERCLOS1, PERCLOS2, std of blink duration & 0.8436 & 0.0940 \\
		& SFBS & 175 & blink frequency, reopening duration, amplitude-velocity ratio of the opening phase, PERCLOS1, PERCLOS2, eyelid cleft, head nodding, head bobbing, std of blink duration, std of mean area defined by the blink, std of PERCLOS2, amplitude-velocity ratio of the closing phase & 0.8419 & 0.0890 \\ 		
		multiclass & SFS & 200 & std of blink duration, eyelid cleft, PERCLOS2, head bobbing, mean area defined by the blink, head nodding, std of PERCLOS2, std of opening duration & 0.6884 & 0.0543 \\
		& SFFS & 175 & head nodding, eyelid cleft, reopening duration, blink frequency, head bobbing, std of reopening duration, PERCLOS2, blink duration, std of average opening velocity & 0.6953 & 0.0768 \\
		& SFBS & 50 & blink duration, opening duration, reopening duration, PERCLOS2, eyelid cleft, head nodding, head bobbing, std of reopening duration, std of average opening velocity, PERCLOS1 & 0.6923 & 0.0801 \\
		\hline \hline
	\end{tabular}
	\caption{Summary of the best performing models resulting from different feature
		selection approaches}\label{tab:eval_FS}
\end{table*}

\subsection{Model Refinement}
The results above reveal promising classification models which yield fair performances on the given dataset (especially in the binary classification setting). As there is still room for improvement, some attempts were made in this work to further raise the BA's and reduce the FDR's. \par

Until now, the classification threshold for the binary classification was set to 0.5, which equals a majority voting within the neighborhood of size $k$. Analyzing the Area Under the Receiver Operating Characteristics (AUROC) and adapting this classification threshold aimed for a further performance boost. Unfortunately, no significant improvements in terms of classification performance were achieved in this work.\par

The results in Table \ref{tab:eval_FS} show that especially the multiclass classification models need further improvements. This can be tackled with different approaches following the 'divide and conquer' principle \cite{Galar2011a}. One of the most promising ones is the so-called One-vs.-One (OvO) technique. This strategy basically splits a $M$-way multiclass classification into one binary classification for each pair of classes. This results in $\frac{M(M-1)}{2}$ binary classifiers, each being responsible for distinguishing between the classes of each pair. The classification itself is done with a majority voting scheme. The advantage of this technique is that dedicated feature sets can be used for each of the classification problems. This increases the class separability and might lead to improved classification results. \par

In this work the three binary classifiers \textit{awake vs. questionable}, \textit{awake vs. drowsy} and \textit{questionable vs. drowsy} are designed independently with the SFS method. Thereby the best feature sets were chosen according to their BA. Furthermore, a weighted voting strategy is applied, where the weight for each vote is given by the BA of the classifier itself \cite{Galar2011a}. As an additional advantage this method prevents a tie between the classes in case all of them receive the same amount of votes. \par

Table \ref{tab:ovo} presents the binary classifiers for the OvO classification identified by the SFS algorithm. \par 
\begin{table*}[t]
	\centering
	\begin{tabular}{p{2cm}p{0.5cm}p{12.5cm}p{0.75cm}}
		\hline \hline classifier & $k$ & feature set & BA \\ \hline
		\textit{awake vs. questionable} & 325 & eyelid cleft, reopening duration, head bobbing, amplitude-velocity ratio of the closing phase, PERCLOS1 & 0.7912 \\
		\textit{awake vs. drowsy} & 75 & reopening duration, head bobbing, eyelid cleft, mean area defined by the blink, amplitude-velocity ratio of the closing phase, PERCLOS1, blink duration, PERCLOS2, blink frequency, std of the opening duration & 0.9305 \\
		\textit{questionable vs. drowsy} & 125 & std of the closing duration, PERCLOS2, std of the blink duration, head bobbing, blink duration, PERCLOS1 & 0.7664 \\ 		
		\hline \hline
	\end{tabular}
	\caption{Details on the binary classifiers for the OvO classification}\label{tab:ovo}
\end{table*}
Although the individual binary classifiers exhibit improved classification performances, a BA of 0.7002 was achieved by combining the classifiers. This result is slightly better compared to the regular multiclass classification, due to the loss of the gained classification sensitivities when merging the classifiers. \par

Table \ref{tab:eval_FS} also shows that some features, namely PERCLOS2, eyelid cleft, head nodding and head bobbing are included into the best feature sets for the multiclass classification. The OvO technique allows for analyzing which features are sensitive to which class. For example, features PERCLOS1 and head bobbing seem to divide all classes from each other, as they are included in every feature subset of the three classifiers \textit{awake vs. questionable}, \textit{awake vs. drowsy} and \textit{questionable vs. drowsy}. Apparently, the classifiers use the reopening duration and the eyelid cleft to distinguish the \textit{awake} state from the other two. This explains their importance and justifies why they are selected in almost all feature sets in the first place.

\section{Discussion} \label{sec:discussion}
The outcomes of this work provide interesting insights into the effect of drowsiness on a driver’s blink properties and head movements and their usefulness for drowsiness classification. In this Chapter the outcomes will be discussed with respect to their contribution towards the aims defined earlier.\par

The foundation of a high-performing $k$-NN model is the correct choice of its parameter $k$ and the features. High-performing feature subsets were identified with the help of several feature selection techniques (wrapper methods). Different algorithms with their respective strengths and weaknesses were tested. The resulting feature subsets enabled deep insights into the connection between the blink behavior, the head movements and the drowsiness levels. The experiments showed that in most cases more compact models (lower $k$, small feature subset) achieve high classification performances whereas overly complex models exhibit lower ones. \par

The multiclass classification posed a particular challenge. With the given model design, the wrapper methods were not able to set up an efficient and high-performing model for the regular multiclass classification. This indicates that the multiclass classification requires a different model design and setup. This was attempted with a special multiclass classification technique where multiple binary classifiers are constructed to distinguish between the given classes. It can be assumed that the classification performance for the multiclass classification problem could be improved by optimizing multiple individual classifiers for subproblems. The combination of the individual classifiers achieved marginally improved classification results compared to the regular multiclass classification approach. The authors of \cite{Galar2011a} mentioned that even if no significant differences are found, the OvO technique is usually more robust. Besides, the OvO approach allows for more insights in the details and properties of the driver drowsiness state classification, specifically about the separation of the three classes \textit{awake}, \textit{questionable} and \textit{drowsy}. 

\section{Conclusion} \label{sec:conclusion}
The aim of this work was to estimate the driver's state by extending the driver drowsiness detection in vehicles using signals of a driver monitoring camera. We have developed and evaluated a $k$-Nearest Neighbor algorithm for the driver's state classification, with a focus on the selection of suitable features. For this purpose, a sufficiently large dataset was recorded and analyzed. Multiple head movement and blink features were derived from the recorded eye closure signal, serving as the basis for the following model design. A key element of the $k$-NN-based classification was the selection of appropriate features. Our approach achieved a balanced validation accuracy of 84.2\% and 70.0\% in the binary and multiclass classification setting, respectively. Despite some uncovered challenges, the suggested classification method provides valuable insights into the effects of drowsiness on blinking behavior and head movements. Thus, it lays the foundation for the development of a driver drowsiness monitoring system which further increases road safety.  The next step towards such a system is to validate its robustness by applying the findings to real-world data.

\bibliographystyle{IEEEtran}
\bibliography{bibliography}

\end{document}